\documentclass[runningheads]{llncs}
\usepackage{graphicx}
\usepackage{multirow}

\usepackage{color}
\usepackage{multirow}
\usepackage{tabu}
\usepackage{bbm}
\usepackage{diagbox}
\usepackage[english]{babel}
\usepackage{comment}
\usepackage{array}
\usepackage{amstext}
\usepackage{makecell}
\usepackage{caption}
\usepackage{tablefootnote}
\usepackage{babel}
\usepackage{booktabs} 
\usepackage{hyperref}

\usepackage[ruled,vlined]{algorithm2e}
\newcommand{\nosemic}{\renewcommand{\@endalgocfline}{\relax}}% Drop semi-colon ;
\newcommand{\dosemic}{\renewcommand{\@endalgocfline}{\algocf@endline}}% Reinstate semi-colon ;
\usepackage{chngcntr}
\usepackage{placeins}
\usepackage{amsfonts}       % blackboard math symbols
\usepackage{nicefrac} 
\usepackage{amsmath,amssymb} % define this before the line numbering.
\usepackage{amstext}

\newcommand{\ie}{\textit{i.e.}}
\newcommand{\eg}{\textit{e.g.}}
\newcommand{\miap}{\emph{MIA}-Prognosis}

\begin{document}
\title{\miap: A Deep Learning Framework to Predict Therapy Response}
\titlerunning{\miap: A Framework to Predict Therapy Response}

\author{Jiancheng Yang\inst{1,2,3,}\thanks{These authors have contributed equally: Jiancheng Yang and Jiajun Chen.}  \and Jiajun Chen\inst{3,\star} \and Kaiming Kuang\inst{3}\and  \\
Tiancheng Lin\inst{1} \and Junjun He\inst{1}  \and Bingbing Ni\inst{1,2,4,}\thanks{Corresponding author: Bingbing Ni.}}
%index{Yang, Jiancheng} 
%index{Chen, Jiajun} 
%index{Kuang, Kaiming}
%index{Lin, Tiancheng}
%index{He, Junjun}
%index{Ni, Bingbing}

\authorrunning{J. Yang et al.}
% First names are abbreviated in the running head.&
\institute{Shanghai Jiao Tong University, Shanghai, China\\
	\and MoE Key Lab of Artificial Intelligence, AI Institute, Shanghai Jiao Tong University\\
	\and Dianei Technology, Shanghai, China
	\and Huawei Hisilicon, Shanghai, China\\
	\email{jekyll4168@sjtu.edu.cn}	\\
}

\maketitle              % typeset the header of the contribution
\begin{abstract}
	Predicting clinical outcome is remarkably important but challenging. Research efforts have been paid on seeking significant biomarkers associated with the 
	therapy response or/and patient survival. However, these biomarkers are generally costly and invasive, and possibly dissatifactory for novel therapy. On the other hand, multi-modal, heterogeneous, unaligned temporal data is continuously generated in clinical practice. This paper aims at a unified deep learning approach to predict patient prognosis and therapy response, with easily accessible data, \eg, radiographics, laboratory and clinical information. Prior arts focus on modeling single data modality, or ignore the temporal changes. Importantly, the clinical time series is asynchronous in practice, \ie, recorded with irregular intervals. In this study, we formalize the prognosis modeling as a \textbf{multi-modal asynchronous} time series classification task, and propose a \miap~framework with \textbf{Measurement}, \textbf{Intervention} and \textbf{Assessment} (MIA) information to predict therapy response, where a Simple Temporal Attention (SimTA) module is developed to process the asynchronous time series. Experiments on synthetic dataset validate the superiory of SimTA over standard RNN-based approaches. Furthermore, we experiment the proposed method on an in-house, retrospective dataset of real-world non-small cell lung cancer patients under anti-PD-1 immunotherapy. The proposed method achieves promising performance on predicting the immunotherapy response. Notably, our predictive model could further stratify low-risk and high-risk patients in terms of long-term survival. A reference implementation in PyTorch is open source at \url{https://github.com/M3DV/SimTA}.

\end{abstract}

\section{Introduction}

Modeling patient prognosis is a challenging but important topic in clinical research, where researchers analyze and predict clinical outcomes including response to certain therapy (\eg, radiotherapy, chemotherapy, surgery, immunotherapy for oncology), patient progression-free survival (PFS) and overall survival (OS). Research efforts have been paid on seeking significant biomarkers, \eg, EGFR mutation for EGFR-TKI therapy \cite{Yu2013AnalysisOT}, 
PD-L1 expression and tumor mutational burden (TMB) for immunotherapy \cite{Gibney2016PredictiveBF}. However, these biomarkers are generally costly and invasive, and possibly dissatisfactory for novel therapy, \eg, anti-PD-1 and anti-PD-L1 immunotherapy \cite{Sacher2016BiomarkersFT}. With more novel revolutionary therapy (including combination therapy \cite{Jain2001NormalizingTV}) available, a unified analytic framework for modeling patient prognosis is urged.

We address this issue via emerging deep learning technology by mining clinical data, \eg, electronic health records (EHR) \cite{Rajkomar2018ScalableAA}. Specifically, we focus on a unified approach to model patient prognosis under certain therapy. Prior arts are generally developed on a single data modality \cite{Hosny2018DeepLF,Sun2018ARA}. Besides, only a few studies \cite{Xu2019DeepLP} take into account the temporal / serial information. In clinical practice, multi-modal temporal data is continuously generated with numerous kinds of sensors and records. It is remarkably valuable to mine the easily accessible information to develop the prognosis prediction system, \eg, radiographics, laboratory and clinical information. We formalize the prognosis modeling as a \textbf{multi-modal asynchronous} time series classification task, and propose a \miap~framework with \textbf{Measurement}, \textbf{Intervention} and \textbf{Assessment} (MIA) information, where Measurement and Intervention information are treated as inputs of multi-modal asynchronous time series to predict the Assessment as ground truth (details in Sec.~\ref{sec:framework}). 

An algorithmic challenge is how to effectively and efficiently process multi-modal asynchronous time series like clinical information. Binkowski \emph{et al.} \cite{Binkowski2017AutoregressiveCN} propose a gated CNN for asynchronous time series analysis, where asynchronous time intervals are regarded as input features. This approach might not be suitable for the clinical scenario since it is not essentially asynchronous; data is needed to learn representation for time intervals. What we need for real-world clinical data processing is a natively asynchronous model, which is flexible and light-weight to learn from expensive clinical data. Inspired by recent advances in natural language processing (NLP), \eg, attention transformers \cite{Vaswani2017AttentionIA,yang2019modeling,Devlin2019BERTPO} and relative position encoding \cite{Shaw2018SelfAttentionWR}, we propose a Simple Temporal Attention (SimTA) module to process asynchronous time series, where attention matrix is learned simply from the time intervals of asynchronous time series (details in Sec.~\ref{sec:simta}). 

The SimTA module is proven to be superior to standard RNN-based approaches in a synthetic asynchronous time series prediction dataset. Moreover, we experiment the proposed \miap~framework on an in-house retrospective dataset of real-world non-small cell lung cancer (NSCLC) under anti-PD-1 immunotherapy. Our predictive model achieves promising performance on predicting the immunotherapy response after 90 days. Notably, this model could further stratify low-risk and high-risk patients in terms of long-term survival.

\section{Methods}

\subsection{MIA Prognosis: The Framework} \label{sec:framework}

\paragraph{\textbf{Categorizing Clinical Information.}}

	In clinical practice, data of a numerous variety of modalities is collected. Most medical data is unaligned in time steps, which means that it has varying intervals between adjacent steps in time series. Such limitations call for a unified framework that integrates asynchronous data of different modalities. To address this issue, we first divide clinical information into three categories according to data sources: \textbf{measurement}, \textbf{intervention} and \textbf{assessment}, which defined our \miap~framework. Measurement data comes from medical examinations such as imaging data (computed tomography, ultrasound, X-ray), laboratory and genetic tests. Measurement is the main information in our \miap~framework. Interventions include actions such as injections and operations. Assessment evaluates the effectiveness of interventions, \eg, Response Evaluation Criteria in Solid Tumors (RECIST) \cite{Eisenhauer2009NewRE}, or 1-year overall survival rates. In this study, we use RECIST to obtain the ``ground truth''\footnote{RECIST is not theoretecally perfect. We refer to ``ground truth'' in a clinical sense.} of therapy response, where complete response (CR), partial response (PR), stable disease (SD) are regarded as response (R), and PD (progressive disease) is regarded as non-response (non-R). Note that measurement, intervention and assessment can also be categoried into either serial or static data, which depends on its status over time. These categorizations are the basis of our framework's capability of integrating heterogeneous multi-modal data.

\paragraph{\textbf{Model Overview.}}

\begin{figure}[tb]
	\includegraphics[width=\textwidth]{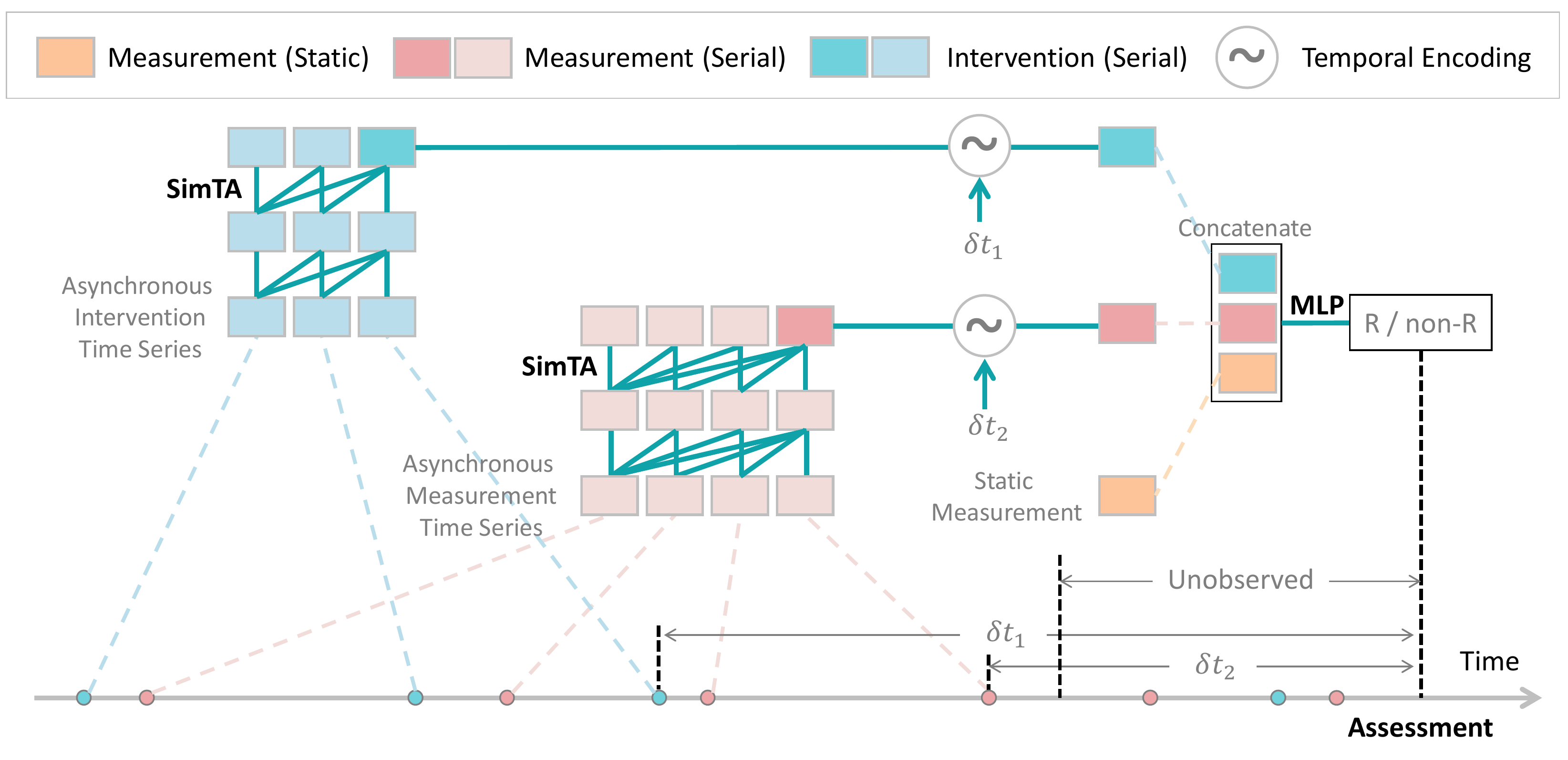}
	\caption{The \miap~framework, with Measurement, Intervention and Assessment information. The asynchronous time series is encoded by the proposed Simple Temporal Attention (SimTA) module into a summary vector. The summary vector is further added with a temporal encoding of time intervals between the assessment time and the last time stamp ($\delta t_1$ and $\delta t_2$). Together with static information, these features predict the therapy response (R / non-R) after an unobserved period.} \label{fig:framework}
\end{figure}

	We propose a framework that integrates multi-modal data in asynchronous time series, named \miap. Fig.~\ref{fig:framework} gives an overall description of our framework. Due to the fact that clinical data of different modalities is usually unaligned in time, it is impractical to simply concatenate these vectors together and pad zeroes at the time step where a certain modality is missing. Therefore, we process each modality independently in our framework. We pass serial data of each modality through its own SimTA module, which outputs a summary vector. The summary vector is added with a temporal encoding (adapted from position encoding \cite{Vaswani2017AttentionIA}) of time intervals between the assessment time and the last time stamp. Static data goes through a multi-layer perceptron (MLP) that encodes it in high-dimensional embedding. We then concatenate summary vectors of serial data with static embedding, and input the concatenated vector into another MLP to give the final prediction of therapy response (R: response / non-R: non-response) after an unobserved period. 
	
	Existing deep sequential models, such as recurrent neural network (RNN), assume that time series data is synchronous in nature. However, this assumption does not hold in the context of clinical practice. Here we introduce a new module to help us process asynchronous data, named SimTA. Inspired by recent advances in natural language processing (NLP), \eg, attention transformers \cite{Vaswani2017AttentionIA,Devlin2019BERTPO} and relative position encoding \cite{Shaw2018SelfAttentionWR}, SimTA utilizes time interval information of asynchronous series to generate attention matrix, capturing temporal relationships between asynchronous time steps. It is worth noting that the latest steps in time series of different modalities are not likely to coincide with each other. In such cases, we use temporal encoding to make use of this information.

\subsection{Simple Temporal Attention for Asynchronous Time Series} \label{sec:simta}

	Simple Temporal Attention (SimTA) is the key factor that enables our framework to process asynchronous times series. Let $X \in \mathbb{R}^{T \times C}$ be an asynchronous time series of length $T$, and $\boldsymbol{\tau}=[\tau_{1},\tau_{2},\dots,\tau_{T-1}] \in \mathbb{R}^{T-1}$ be the time interval vector between any adjacent time steps. A general formula of a single SimTA module can be described as:
	\begin{equation} 
		SimTA(X,\boldsymbol{\tau})=softmax(A)\sigma(f(X)),
	\end{equation}
where $\sigma$ denotes an activation function of our choice, and $f$ is a fully-connected layer. $A=S(X,\tau) \in \mathbb{R}^{T\times T}$ is the attention matrix that encodes relations between any two time steps. Matrix $A$ can be calculated using any edge-aware attention mechanism, \eg, multi-head self attention \cite{Vaswani2017AttentionIA} with relative position encoding~\cite{Shaw2018SelfAttentionWR}. In this study, we use an extremely simplified version that only encodes linearly time intervals, which is validated effective in our experiments:
	\begin{equation}
	A=S(\boldsymbol{\tau})=
	\begin{bmatrix}
	0,       & -\infty,& -\infty, & \dots & -\infty \\
	-\lambda \tau_1+\beta,     & 0, & -\infty, & \dots & -\infty \\
	-\lambda (\tau_1+\tau_2)+\beta,     & -\lambda \tau_2+\beta,  & 0, & \dots & -\infty\\
	\vdots & \vdots & \vdots & \ddots & \vdots\\
	-\lambda \sum_{1}^{T-1}{\tau_i}+\beta,    &-\lambda \sum_{2}^{T-1}{\tau_i}+\beta,  & -\lambda \sum_{3}^{T-1}{\tau_i}+\beta,  & \dots & 0
	\end{bmatrix},
	\label{eq:simta_attn}
	\end{equation}
	where $\lambda\in \mathbb{R}^+$ and $\beta\in \mathbb{R}$ are trainable parameters that apply a linear transformation on $\boldsymbol{\tau}$. The simplicity of this attention mechanism can help us cope with overfitting as well, considering we only have limited amount of data. The formula is based on the assumption that the more recent time steps should have stronger correlations with the current time step than further ones. Complex temporal information can be captured by stacking multiple SimTA modules. The complete SimTA model pipeline is a SimTA block of one or multiple SimTA modules, which outputs a summary vector, followed by an MLP that outputs the final prediction with softmax activation. To the best of our knowledge, the proposed SimTA is the first study to introduce attention mechanism to asynchronous time series analysis with proven effectiveness.

\subsection{Counterpart Approaches}
	To demonstrate SimTA's capability of learning asynchronous temporal relations, we bring in LSTM (Long Short-term Memory) \cite{Hochreiter1997LongSM,Greff2017LSTMAS} as comparison. LSTM is a special type of RNN designed to capture relations over extended time intervals in sequences. In our experiments, we use LSTM models with a comparable size and the same training configuration as SimTA. Let $X_{i}$ be the value of the time series at time step $t_{i}$. Three LSTM approaches are tested, which differ in their input: (1) Only $X_{i}$; (2) $X_{i}$ and time intervals $\tau_{i}$; (3) $X_{i}$ and time stamps $t_{i}$. We mark them as LSTM, LSTM(i) and LSTM(s), respectively.

\section{Experiments}

\subsection{Proof of Concept on Synthetic Dataset}
	We first validate the superiority of the proposed SimTA over RNN on asynchronous time series using a synthetic dataset.
	\paragraph{\textbf{Dataset and Experiment Settings.}}
	The synthetic dataset consists of the summation of $N$ trigonometric functions with random periods. Each time series is computed as:
	\begin{equation}
	X_{t}=\sum_{j=1}^{N}[{\alpha_{j}\sin(\omega_{j}\pi t+b_{j})+\beta_{j}}]+\eta\epsilon
	\end{equation}
	where $N$ is the number of trigonometric functions involved. $\epsilon$ is a white noise following the standard normal distribution, whose magnitude is controlled by a constant $\eta$. In our experiments, we choose $N=10$ and $\eta=0.5$. We generate 10,000 such asynchronous series, which are split in 80/20 for training/validation, respectively. The training/validation data is sampled once from the predefined distribution and fixed throughout the experiment.
	
\begin{figure}[tb]
	\includegraphics[width=\textwidth]{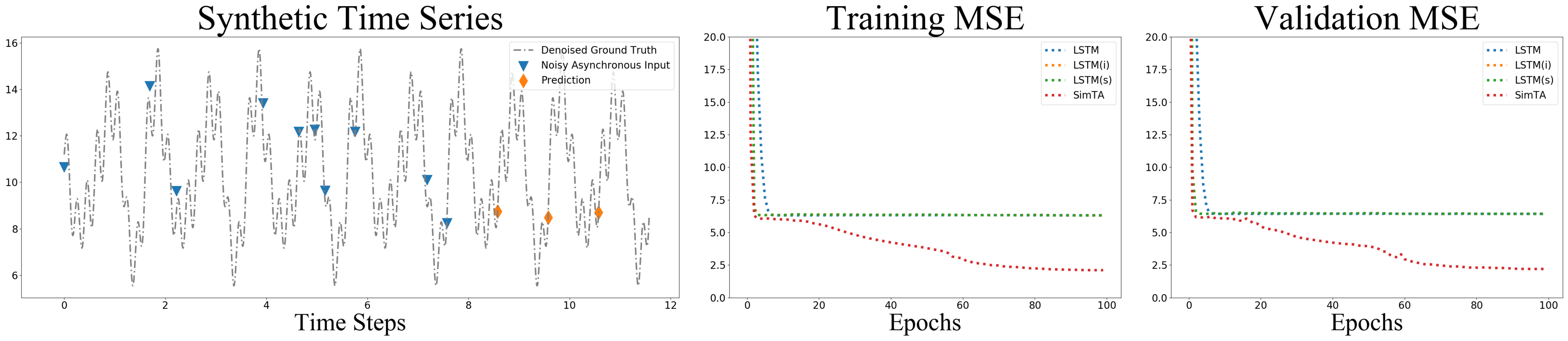}
	\caption{The synthetic time series and MSE loss curves of LSTM, LSTM(i), LSTM(s) and the proposed SimTA. \textbf{Left}: The illustration of a data sample. \textbf{Middle}: Training MSE loss curves. \textbf{Right}: Validation MSE loss curves. We clip the y axis in both loss curves for the sake of better visualization.} \label{fig:poc}

	\end{figure}

	During training, 10 different $X_{t}$ are randomly sampled from each series. The time intervals between any two adjacent points follow a uniform distribution between $0$ and a maximum interval level $I$. The model is tasked to predict the next 3 points ($+1,+2,+3$) following the last one in the input. Fig.~\ref{fig:poc} shows a sample series of the synthetic dataset.
	In our experiments, we compare performances of SimTA and three LSTM models, which are LSTM, LSTM(i) and LSTM(s). For LSTM(i) and LSTM(s), $X_{t}$ and the time information are concatenated into one vector.  SimTA follows the model described in Sec. \ref{sec:simta}.

	\paragraph{\textbf{Results.}}
	We train all four models for 100 epochs. Fig.~\ref{fig:poc} shows the training and validation mean squared error over the training phase. SimTA outperforms all three LSTM models on both training and validation data. It achieves significantly lower MSE (2.197 on SimTA and 6.427 on all LSTM approaches) compared with LSTM. The time information does help LSTM(i) and LSTM(s) to converge faster than vanilla LSTM, but all three end up with errors at the same level. It is worth noting that the LSTM model variants underfit the training set. From the observations above, we conjecture that the proposed SimTA outperforms existing standard sequential models such as LSTM for asynchronous time series.

\subsection{Predicting Response to Anti-PD-1 Immunotherapy for Non-Small Cell Lung Cancer (NSCLC)}

	\paragraph{\textbf{Background.}}

	Lung cancer is the most commonly diagnosed cancer worldwide. According to \cite{Bray2018GlobalCS}, lung cancer accounts for 18.4\% of the global cancer deaths in 2018. NSCLC makes up 80\%-85\% of these cases. Deep learning has shown its potential in precision medicine for lung cancer \cite{zhao20183d,zhao2019toward,yang2019development,yang2020relational}. Recently, immunotherapy has been proven to remarkably increase the overall survival and the life quality of patients with a variaty of cancers, including NSCLC. However, only a small percentage of patients benefit from immunotherapy and show lasting responses. There has been research on the identification of response predictors, whereas most of the effort are focused on biopsy analyses and serum biomarkers, \eg, PD-L1 expression and tumor mutation burden (TMB) for first-line immunotherapy. These methods are expensive, invasive, and not always consistently associated with tumor responses. Furthermore, no biomarker is available for predicting second-line NSCLC immunotherapy outcome so far . Such limitations emphasize the necessity for convenient, economical and non-invasive indicators, especially for second-line immunotherapy treatment.

\paragraph{\textbf{Dataset and Experiment Settings.}}

In this retrospective study, 99 patients with advanced or metastatic stage IIIB and IV NSCLC under second-line immunotherapy are included. The dataset includes 793 CT scans, 1335 laboratory blood tests, 99 clinical data, and 320 response evaluations as per RECIST1.1 \cite{Eisenhauer2009NewRE}. All data is further categorized into serial data and static data.

The CT scans are labelled by one radiologist with 8 years of experience, by manually segmenting the volume of interest (VOI) of the target lesion in each scan. An oncologist with 30 years of experience reviewed and confirmed the segmentation. CT volumes and segmentation masks are resampled to uniform spacing ($1mm \times 1mm \times 1mm$), with B-spline interpolation for CT volumes and nearest-neighbor interpolation for VOI masks. We use radiomics features~\cite{Gillies2016RadiomicsIA} to represent the radiological features due to limited number of samples. With large data available, a fully end-to-end CNN could also be used as the feature extractor. 107 radiomics features are extracted from each VOI using PyRadiomics~\cite{Griethuysen2017ComputationalRS}. Radiomics features are treated as serial data unless there is only one CT examination. The serial blood test features are in 22 dimensions and static clinical information features are in 18 dimensions. All categorical features are encoded in one-hot vectors. Numeric features are normalized by removing the mean and scaling to unit variance to ensure stable training and faster convergence. Intervention information is one-hot encoded (\ie, a binary flag at a time step). Serial radiomics, laboratory blood test and intervention is asynchronous in time.

In our experiments, models are tasked to output binary predictions of R (response) or Non-R (non-response) of each response evaluation, with static data and all serial data before 90 days prior to time of response assessment. We use binary cross entropy (BCE) as the loss function. SimTA model is optimized with Adam optimizer~\cite{Kingma2014AdamAM}. We split the 99 patients into 3-fold (33 patients in each fold), and perform 3-fold cross validation for evaluating our method. Hyperparameters of model structure and training configuration are chosen using a grid search with bootstrap on the training dataset in each cross validation fold. To verify the effectiveness of SimTA on asynchronous time series, we include LSTM, LSTM(i) and LSTM(s) in our ablation study as comparison. LSTM models of comparable parameters are trained under similar setting.

We further associate model prediction with clinical survival benefits, specifically, overall survival (OS) and progression-free survival (PFS). A cutoff value of 0.5 is used for stratifying patients into high-risk and low-risk groups. The serial data before 90 days prior to time of response assessment is used as input to the trained model. Kaplan-Meier analysis and log-rank test for the survival analysis validate the effectiveness of our method in terms of patient survival. 

\paragraph{\textbf{Results.}}

\begin{table}[tb]
	\centering
	\caption{Model performance of on predicting immunotherapy response, including standard RNN approaches (LSTM, LSTM(i), LSTM(s)) instead of the proposed SimTA, and our methods with multi-modal and single-modal inputs.} \label{tab:model-performance}
	\begin{tabular*}{\hsize}{@{}@{\extracolsep{\fill}}l|ccc|ccc@{}}
		\toprule
		Methods & LSTM  & LSTM(i) & LSTM(s) & Ours & Ours w/o radiomics & Ours w/o lab \\
		\midrule
		AUC  & 0.71&0.71&0.70 & 0.80&0.47&0.58 \\
		\bottomrule
	\end{tabular*}

\end{table}

\begin{figure}[tb]
	\includegraphics[width=\textwidth]{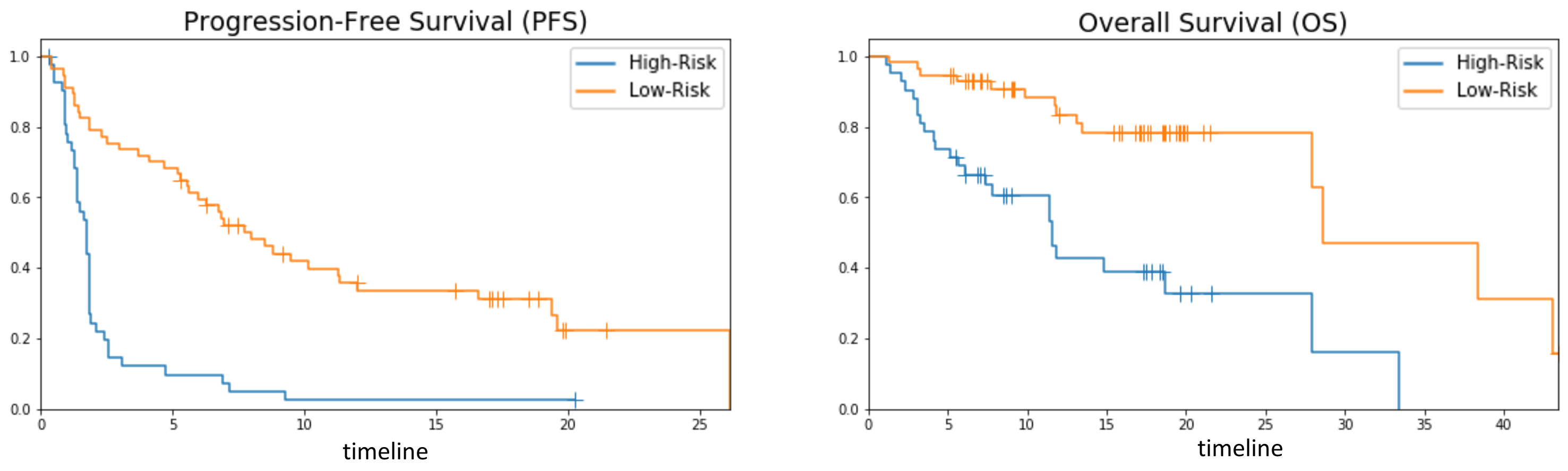}
	\caption{Model performance on predicting NSCLC patient survival under anti-PD-1 immunotherapy. \textbf{Left}: Patient survival curve visualized by Kaplan-Meier (K-M) plot of progression-free survival (PFS), p-value of log-rank test to high/low-risk groups is $<0.01$. \textbf{Right}: K-M plot of overall survival (OS), with $p<0.01$.} \label{fig:km_plot}

\end{figure}

As depicted in Table \ref{tab:model-performance}, promising results are observed in predicting immunotherapy outcome using the proposed framework. The area under curve (AUC) of receiver operating characteristic (ROC) curve is 0.80 with our \miap~framework, whereas vanilla LSTM, LSTM(i) and LSTM(s) are achieving 0.71, 0.71 and 0.70 AUC respectively. The LSTM counterparts does not totally fail in this case because the patients are taking CT scans and blood tests on a fairly regular schedule, the interval variance is mostly smaller than seven days. Still, SimTA outperforms LSTM by a large margin in this “mildly” asynchronous serial data modelling task. We also validate the necessity of multi-modal input. Without radiomics feature or laboratory blood test results, our framework reaches very low AUC of 0.47 and 0.58 respectively, suggesting the significance of multi-modal model in this task. Moreover, as shown in Fig. \ref{fig:km_plot}, the p-values for Kaplan-Meier analysis are significant in both PFS and OS tests. Therefore, our predictive model could further stratify the low- and high-risk patients in terms of patient survival.

\section{Conclusion and Further Work}

In this paper, we focus on a unified deep learning framework to predict therapy response, with easily accessible clinical data. The proposed framework named \miap~utilizes clinical information including Measurement, Intervention and Assessment to model patient prognosis. We also propose a Simple Temporal Attention (SimTA) module to process the asynchronous time series. The proof-of-concept experiments validate the superiority of SimTA over standard RNN approaches in asynchronous time series analysis. Moreover, our method is proven effective on an in-house dataset on predicting response to anti-PD-1 immunotherapy for real-world non-small cell lung cancer (NSCLC) patients. Importantly, our predictive model is associated with long-term patient survival in terms of progression-free survival (PFS) and overall survival (OS).

In future studies, it is valuable to apply the proposed \miap~framework on other therapy and diseases. On the other hand, it is also important to design efficient and effective non-linear temporal attention module to enhance temporal relation learning of SimTA. Besides, a fully end-to-end model with CNN-based Radiomics \cite{yang2019probabilistic} to encode the signature of radiographic features is worth exploring. Furthermore, it is interesting to explain what the \miap~models from data-driven approaches.

\subsubsection{Acknowledgment.}
This work was supported by National Science Foundation of China (61976137, U1611461). Authors would like to appreciate the Student Innovation Center of SJTU for providing GPUs.

\bibliographystyle{splncs04}
%\bibliography{reference}

\end{document}